%% file: Paper-3730.tex
\documentclass[runningheads]{llncs}
\usepackage[T1]{fontenc}
\usepackage{tabularx}
\usepackage{amsmath}
\usepackage{amssymb} 
\usepackage{multirow} 
\usepackage[table,xcdraw]{xcolor}
\usepackage{amsmath,graphicx}
\usepackage[utf8]{inputenc}
\usepackage[hidelinks]{hyperref}
\usepackage{fontawesome}
\usepackage{amssymb}
\usepackage{pifont}
\begin{document}
\title{MuST: \textbf{Mu}lti-\textbf{S}cale \textbf{T}ransformers for Surgical Phase Recognition}

%
%
\author{Alejandra Pérez\inst{*} \textsuperscript{\faEnvelopeO} \and
Santiago Rodríguez\inst{*} \textsuperscript{\faEnvelopeO}\and
Nicolás Ayobi \inst{\diamond}\and
Nicolás Aparicio \inst{\diamond}\and
Eugénie Dessevres\inst{}\and
Pablo Arbeláez\inst{}}


%
\authorrunning{A. Pérez, S. Rodríguez et al.}
%
\institute{Center for Research and Formation in Artificial Intelligence, Universidad de los Andes, Bogotá, Colombia
\\
\email{\{a.perezr20, s.rodriguezr2\}@uniandes.edu.co}}

\maketitle

\begin{abstract} Phase recognition in surgical videos is crucial for enhancing computer-aided surgical systems as it enables automated understanding of sequential procedural stages. Existing methods often rely on fixed temporal windows for video analysis to identify dynamic surgical phases. Thus, they struggle to simultaneously capture short-, mid-, and long-term information necessary to fully understand complex surgical procedures. To address these issues, we propose Multi-Scale Transformers for Surgical Phase Recognition (MuST), a novel Transformer-based approach that combines a Multi-Term Frame encoder with a Temporal Consistency Module to capture information across multiple temporal scales of a surgical video. Our Multi-Term Frame Encoder computes interdependencies across a hierarchy of temporal scales by sampling sequences at increasing strides around the frame of interest. Furthermore, we employ a long-term Transformer encoder over the frame embeddings to further enhance long-term reasoning. MuST achieves higher performance than previous state-of-the-art methods on three different public benchmarks.

\def\thefootnote{$*$ $\diamond$}\footnotetext{These authors contributed equally to this work}


\keywords{Surgical Workflow Analysis  \and Surgical Phase Recognition \and Long-Term Transformers \and Vision Transformers \and Temporal Multi-Scale}

\end{abstract}

\input{Sections/01_Introduction}

\input{Sections/02_Method}
\input{Sections/03_Experiments}
\input{Sections/04_Conclusions}

\begin{credits}
\subsubsection{\ackname} Alejandra Pérez, Santiago Rodriguez, and Nicolás Ayobi acknowledge the support of the 2022 and 2023 UniAndes-DeepMind Scholarships.

\subsubsection{\discintname}
The authors have no competing interests to declare that are relevant to the content of this article. 
\end{credits}

%
%

\bibliographystyle{splncs04}
\bibliography{Paper-3730}

\end{document}


\title{MuST: Supplementary Material}

%

%
%
%

\maketitle              

\begin{figure}
    \centering
    \includegraphics[width=0.65\linewidth]{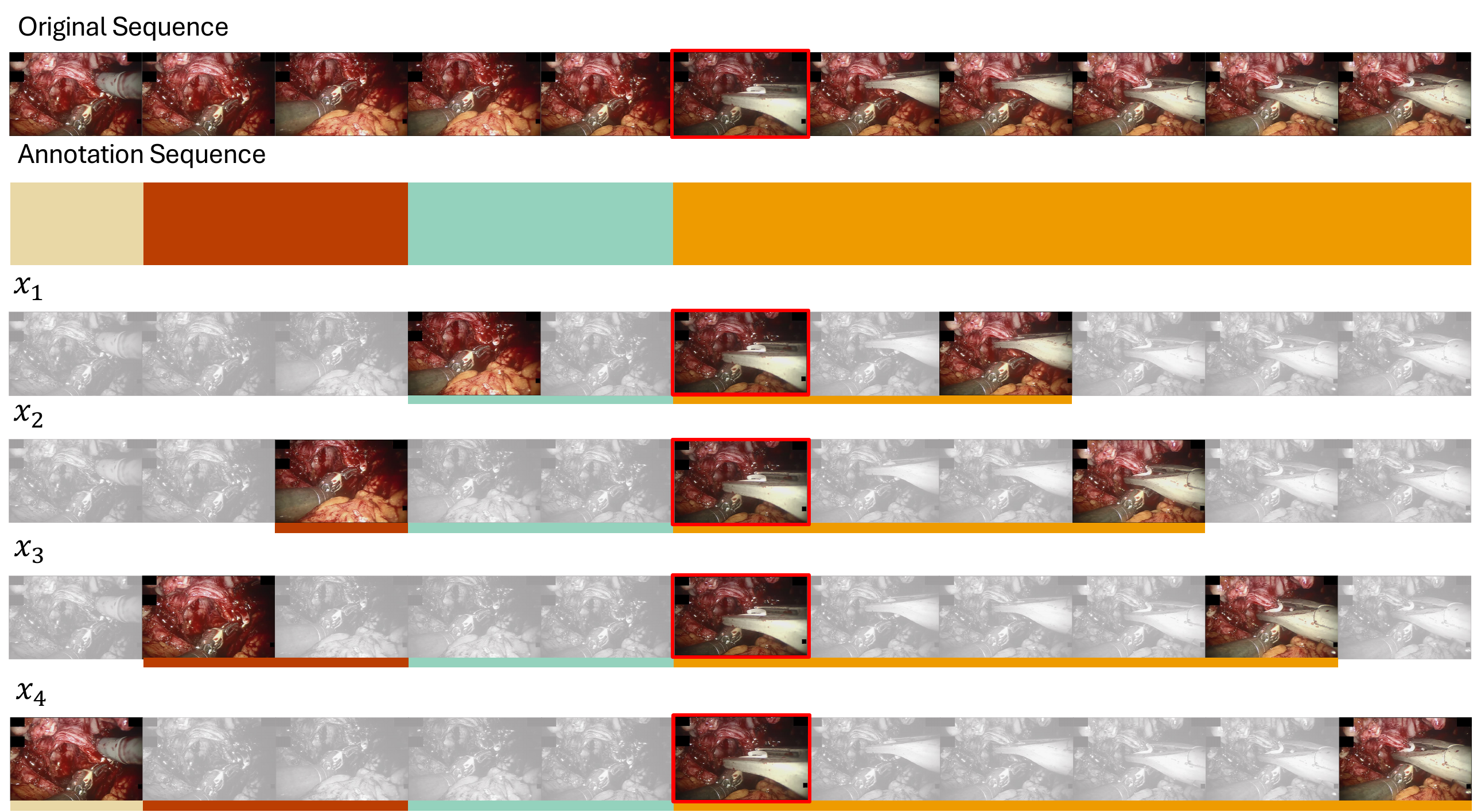}
    \caption{\textbf{Temporal Multi-Scale Pyramid Example.} Illustration of a temporal multi-sequence pyramid centered on the red keyframe. Sampled frames at each level are highlighted. The top levels of the pyramid capture short-term temporal information, while the lower levels comprehend a broader temporal context.}
    \label{fig:temporal_pyramid}
\end{figure}

\begin{figure}
    \centering
    \includegraphics[width=0.8\linewidth]{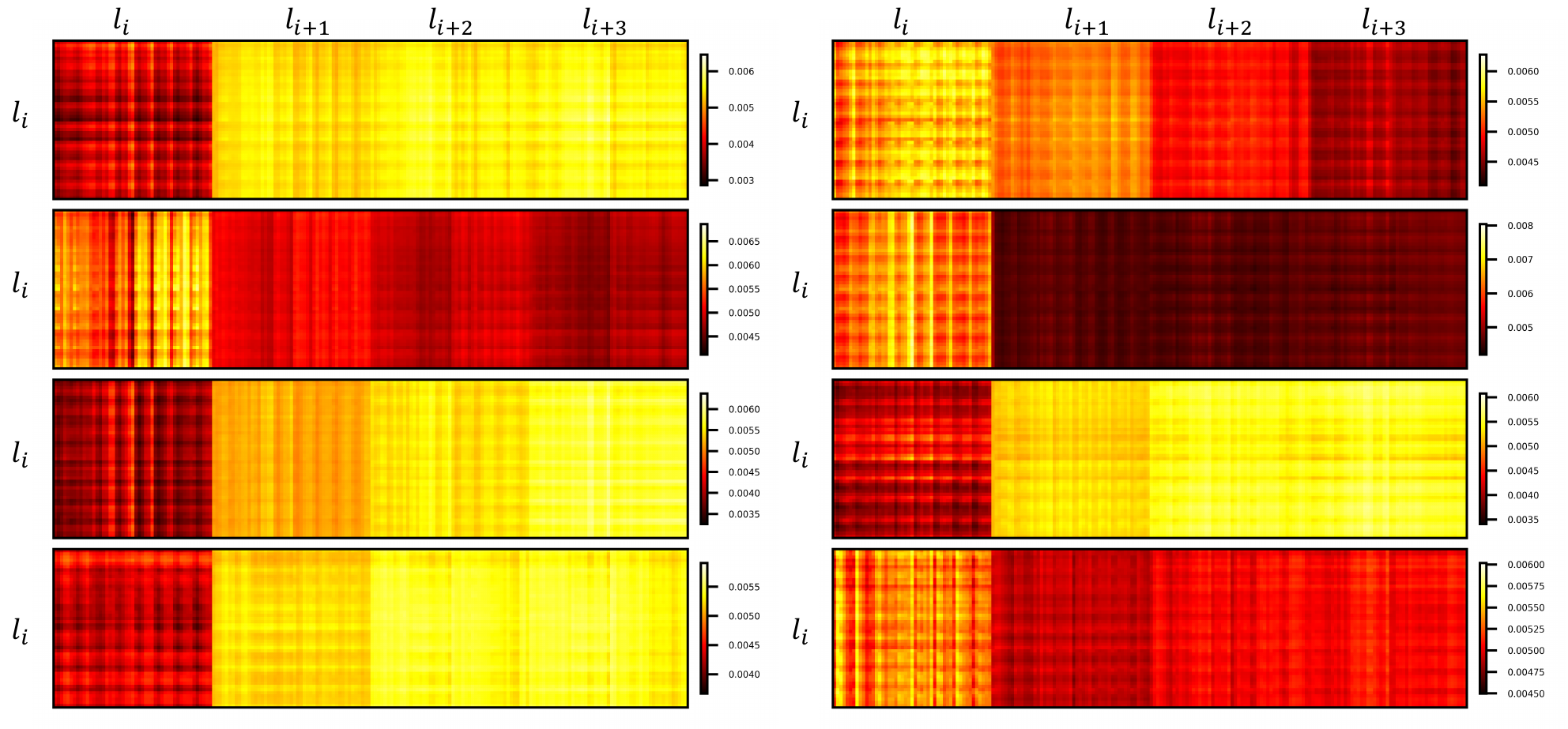}
    \caption{\textbf{Cross-attention maps.} Attention weights from the 8 attention heads of one of the layers of the cross-attention described in Section 2 (Method) for the $l_{i}$ sequence of the set $L$. These attention maps illustrate that each head specializes in capturing different segments of the $L$ set, featuring inter-dependencies across temporal scales.}
    \label{fig:attention_heads}
\end{figure}

\begin{figure}
    \centering
    \includegraphics[width=0.85\linewidth]{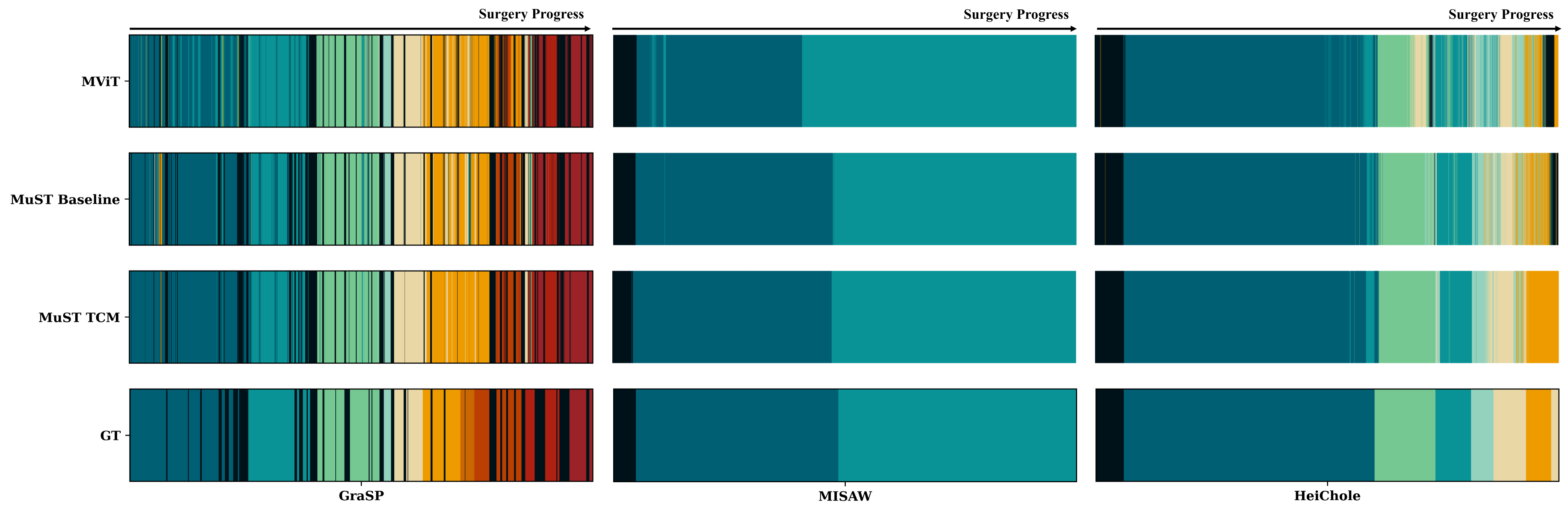}
    \caption{\textbf{MuST visualization performance in every benchmark.} MuST baseline refers to our model without the Temporal Consistency Module (TCM), whereas MuST TCM represents the complete model. Overall visualizations demonstrate that MuST achieves consistent recognition of phases, including reclassification of some segments when using the Temporal Consistency Module.}
    \label{fig:ablation_visuals}
\end{figure}

\begin{table}[]
\centering
\caption{\textbf{Class comparison of MuST with state-of-the-art methods.} MuST baseline refers to our model without the Temporal Consistency Module (TCM), whereas MuST TCM represents the complete model. Consistent results across classes are evident across diverse temporalities within the databases. These results underscore MuST's proficiency in identifying multiple duration phases, attributed to its multi-temporal reasoning. Leveraging information from distinct temporal contexts enhances MuST's predictive capabilities for phases of different durations. The best results are highlighted in bold.}
\label{tab:supp-per_class}
\resizebox{12cm}{!}{%
\begin{tabular}{ccccccc}
\multicolumn{7}{c}{\textbf{GraSP}}                                                                                                                               \\ \hline
Phase      & \multicolumn{1}{c|}{Duration}        & TAPIS         & TeCNO         & \multicolumn{1}{c|}{Trans-SVNet}   & MuST Baseline (Ours) & MuST TCM (Ours) \\ \hline
0        & \multicolumn{1}{c|}{50.60 ±107.06}   & 85.5         & 77.9          & \multicolumn{1}{c|}{74.7}          & 86.8        & \textbf{87.1}            \\
1        & \multicolumn{1}{c|}{285.28±318.04}   & 86.0          & 95.8 & \multicolumn{1}{c|}{94.2}          & 95.6                 & \textbf{97.7}           \\
2        & \multicolumn{1}{c|}{221.55±256.20}   & 76.0          & 87.7          & \multicolumn{1}{c|}{84.5}          & 91.6        & \textbf{93.0}            \\
3     & \multicolumn{1}{c|}{143.33±159.81}   & 83.0          & 83.2          & \multicolumn{1}{c|}{85.3}          & 91.9        & \textbf{94.2}            \\
4      & \multicolumn{1}{c|}{84.73±55.18}     & 83.8          & 67.1          & \multicolumn{1}{c|}{76.1}          & \textbf{90.8}        & 89.5            \\
5 & \multicolumn{1}{c|}{149.16±149.92}   & 85.0          & 78.5          & \multicolumn{1}{c|}{82.3}          & \textbf{89.5}                 & 87.8   \\
6     & \multicolumn{1}{c|}{222.21±264.64}   & 73.5          & 81.6          & \multicolumn{1}{c|}{79.4}          & 79.4                 & \textbf{82.1}   \\
7 & \multicolumn{1}{c|}{108.00±56.66}    & 8.6           & \textbf{68.1} & \multicolumn{1}{c|}{61.1}          & 16.6                 & 22.6            \\
8     & \multicolumn{1}{c|}{74.89±76.06}     & 71.6 & 70.1          & \multicolumn{1}{c|}{66.1}          & 71.8                 & \textbf{76.2}            \\
9    & \multicolumn{1}{c|}{107.00±77.22}    & 59.3          & \textbf{70.2}          & \multicolumn{1}{c|}{67.9}          & 63.8                 & 60.4   \\
10     & \multicolumn{1}{c|}{70.48±80.06}     & 76.3          & 67.8          & \multicolumn{1}{c|}{70.2}          & 78.9                 & \textbf{79.8}   \\ \hline
            &                                      &               &               &                                    &                      &                 \\

\multicolumn{7}{c}{\textbf{MISAW}}                                                                                                                                        \\ \hline
Phase       & \multicolumn{1}{c|}{Duration}        & TAPIS         & TeCNO         & \multicolumn{1}{c|}{Trans-SVNet}   & MuST (Ours)          & MuST TCM (Ours) \\ \hline
0           & \multicolumn{1}{c|}{114.70±62.62}    & 92.1          & 87.8          & \multicolumn{1}{c|}{75.7}          & 94.2                 & \textbf{95.0}   \\
1           & \multicolumn{1}{c|}{2318.22±931.10}  & 97.9          & 99.4          & \multicolumn{1}{c|}{96.6}          & 98.9                 & \textbf{99.5}   \\
2           & \multicolumn{1}{c|}{3651.41±2759.67} & 98.9          & 99.6          & \multicolumn{1}{c|}{98.8}          & 99.4                 & \textbf{99.8}   \\ \hline
            &                                      &               &               &                                    &                      &                 \\
\multicolumn{7}{c}{\textbf{HeiChole}}                                                                                                                                     \\ \hline
Phase       & \multicolumn{1}{c|}{Duration}        & TAPIS         & TeCNO         & \multicolumn{1}{c|}{Trans-SVNet}   & MuST (Ours)          & MuST TCM (Ours) \\ \hline
0           & \multicolumn{1}{c|}{170.38±63.18}    & 86.4          & 89.3 & \multicolumn{1}{c|}{92.3}          & 96.2                 & \textbf{97.6}          \\
1           & \multicolumn{1}{c|}{618.67±461.98}   & 83.0          & 81.9 & \multicolumn{1}{c|}{84.2}          & \textbf{92.3}                 & 91.4            \\
2           & \multicolumn{1}{c|}{127.85±61.96}    & 46.4          & 37.8          & \multicolumn{1}{c|}{40.3}          & \textbf{61.6}                 & 58.0   \\
3           & \multicolumn{1}{c|}{384.68±222.82}   & 73.8          & 79.2          & \multicolumn{1}{c|}{78.1} & 76.6                & \textbf{77.6}          \\
4           & \multicolumn{1}{c|}{86.33±37.20}     & 72.1 & 64.1          & \multicolumn{1}{c|}{\textbf{75.4}}          & 67.2                 & 68.5            \\
5           & \multicolumn{1}{c|}{279.48±208.76}   & 58.5         & 65.6          & \multicolumn{1}{c|}{67.7} & 70.0              & \textbf{73.2}            \\
6           & \multicolumn{1}{c|}{70.12±63.01}     & 64.7          & 69.3 & \multicolumn{1}{c|}{64.9}          & 67.9                 & \textbf{73.0}            \\ \hline
\end{tabular}%
}
\end{table}

%% file: Sections/01_Introduction.tex
\section{Introduction} \label{sec:introduction}

Surgical workflow analysis is critical for computer-assisted surgery, as it aims to understand the operational sequence of stages in surgical procedures \cite{mair2022surgical,kirtac2022surgical,padoy2012statistical}. Henceforth, equipping computer-aided systems with the ability to recognize these workflows would improve automated assistance to surgical teams, facilitate postoperative analysis, and contribute to medical personnel training \cite{wagner2023comparative}. An imperative step towards automatic surgical workflow analysis is identifying the succession of distinct phases throughout a surgical procedure \cite{demir2023review}. Despite advancements in addressing this task, phase recognition remains challenging as it demands adaptable models capable of understanding the inherent variability in phase durations and the high semantic similarity between distinct phases present in surgical data \cite{kirtac2022surgical}. Thus, these models need to incorporate long-term temporal context and ensure precise temporal consistency to achieve a reliable understanding of surgical progressions \cite{liu2023skit}.

Pioneering works for surgical phase recognition used manually designed features to train classical machine learning classifiers \cite{padoy2012statistical,padoy2019machine}. However, Deep Learning methods gradually replaced them due to their superior capacity to represent temporal and spatial contexts. EndoNet \cite{twinanda2016endonet} was the first approach that employed Convolutional Neural Networks (CNNs) to extract features that captured spatial dependencies but did not incorporate the temporal information from surgical videos. To address this limitation, PhaseNet \cite{twinanda2016single}, EndoLSTM \cite{twinanda2017vision},  SV-RCNet \cite{jin2017sv}, and OHFM \cite{yi2019hard} used long short-term memory (LSTM) networks to model temporal dependencies. Nevertheless, LSTM-based methods have limited long-term processing capacities due to their sequential nature and vanishing gradient issues. To mitigate this problem, TMRNet \cite{jin2021temporal} proposed multi-scale non-local operations, but they needed to aggregate global information collaboratively. On the contrary, TeCNO \cite{czempiel2020tecno} used Temporal Convolutional Networks (TCN) with dilated convolutions to achieve long-term temporal context, which limited its comprehension of short-term and fine-grained information. 

\begin{figure}[t]
    \centering
    \includegraphics[width=0.9\columnwidth]{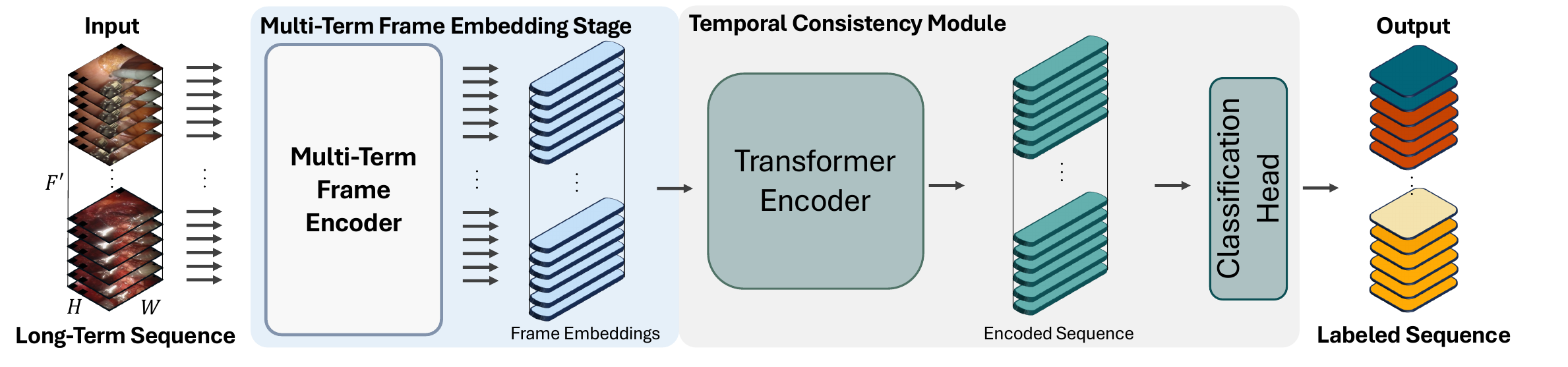}
    \caption{\textbf{MuST} employs a Multi-Term Frame Encoder to generate rich embeddings containing short- and mid-term dependencies for a long-term sequence of $F'$ frames. The Temporal Consistency Module introduces long-term analysis by processing relationships among frame embeddings to enable coherence in the predictions.
    }
    \label{fig:arch}
\end{figure}

Transformers \cite{vaswani2017attention} have proved to be remarkably powerful for sequential data and also for image and video analysis \cite{carion2020end,fan2021mvit,zhu2020deformable,dosovitskiy2020image} due to their ability to model long-term dependencies and extract rich features through self-attention mechanisms. Multiple surgical workflow analysis models like OperA \cite{czempiel2021opera}, SAHC \cite{ding2022exploring}, and Trans-SVNet \cite{gao2021trans} incorporated Transformer layers to TCNs in order to efficiently combine the spatial and temporal features. Nonetheless, their dependence on TCN modeling leads to a loss of finer-grained information, and using temporal-agnostic backbones limits frame embeddings to capture only spatial information. In contrast, TAPIR \cite{valderrama2022towards} and TAPIS \cite{ayobi2024pixelwise} proved the utility of a fully Transformer-based model for multiple surgical workflow analysis tasks and improving frame representation by extracting rich spatio-temporal embeddings. Still, these models process short-term fixed-size windows and employ a per-frame prediction approach that limits its temporal consistency. Recently, SKiT \cite{liu2023skit} achieved long-term reasoning by using critical pooling to record relevant past information, but it also employs a backbone that only extracts spatial dependencies.

Several methods have studied the potential of multi-scale analysis to improve temporal comprehension in generic human activity recognition \cite{slowfast,stergiou2023tempr,girdhar2021anticipative,meng2020ar}. Initially, SlowFast \cite{slowfast} introduced the concept of sub-sampling video sequences at different rates to consider multiple temporal coverages. More recently, TemPr \cite{stergiou2023tempr} extended this approach by using a set of Transformers on various sequences of increasing temporal coverages, enabling information aggregation from diverse scales. Still, these concepts need to be fully explored in surgical workflow analysis. Integrating them could be highly advantageous, as it introduces dynamic window designs for phase recognition that adapt to phases of varying lengths.

This work presents a novel approach to augment flexibility over surgical phase variability by processing multiple temporal scales. For this purpose, we introduce Multi-Scale Transformers for Surgical Phase Recognition (MuST), a fully Transformer-based model composed of two processing stages. The first stage corresponds to a Multi-Term Frame Encoder (MTFE) that captures information from short- and mid-term temporal scales into multi-term frame embeddings. Within this stage, we create a temporal pyramid comprising multiple sequences sampled at progressively increasing rates \cite{stergiou2023tempr}. Inspired by multi-scale image processing Transformers \cite{chen2021crossvit}, we introduce a Multi-Temporal Attention Module that correlates information from each sequence and combines patterns across short- and medium-term periods, offering a flexible understanding of the information wrapping each frame. The second stage is a Temporal Consistency Module (TCM) that employs a lightweight and long-term Transformer encoder to process extensive sequences of multi-term frame embeddings. This approach enables our TCM to encode long-term dependencies within the surgical procedure and predict temporally coherent surgical phase segments. Hence, the  MTFE and the TCM complement each other to achieve holistic temporal modeling of complex surgical procedures.

To sum up, our main contributions are: (1) We introduce a multi-sequence pyramid and a Multi-Temporal Attention Module for surgical phase recognition, allowing the combined analysis of temporal windows with a hierarchical temporal coverage over a surgical video, and (2) we propose a temporal consistency module that leverages Transformers' attention over the extracted frame embeddings to improve the consistency along wide-range temporal segments. 

We demonstrate MuST's superiority over previous state-of-the-art models by extensively evaluating on three public benchmarks. To ensure reproducibility, all our source code and pretrained models are available at \url{https://github.com/BCV-Uniandes/MuST}.

%% file: Sections/02_Method.tex
\section{Method} \label{sec:method}

We present \textbf{Mu}lti-\textbf{S}cale \textbf{T}ransformers for Surgical Phase Recognition (MuST), a two-stage Transformer-based architecture designed to enhance the modeling of short-, mid-, and long-term information within surgical phases. Our method employs a frame encoder that leverages multi-scale surgical context across different temporal dimensions. The frame encoder considers diverse time spans around a specific frame of interest, which we call a keyframe. The keyframe serves as the specific frame that we encode. We construct temporal windows around this keyframe to provide the necessary temporal context for accurate phase classification. Our encoder generates rich embeddings that capture short- and mid-term dependencies. To further enhance long-term understanding, we employ a Temporal Consistency Module (TCM) that establishes relationships among frame embeddings within an extensive temporal window, ensuring coherent phase recognition within an extensive temporal window.

\begin{figure}[t]
    \centering
    \includegraphics[width=0.9\columnwidth]{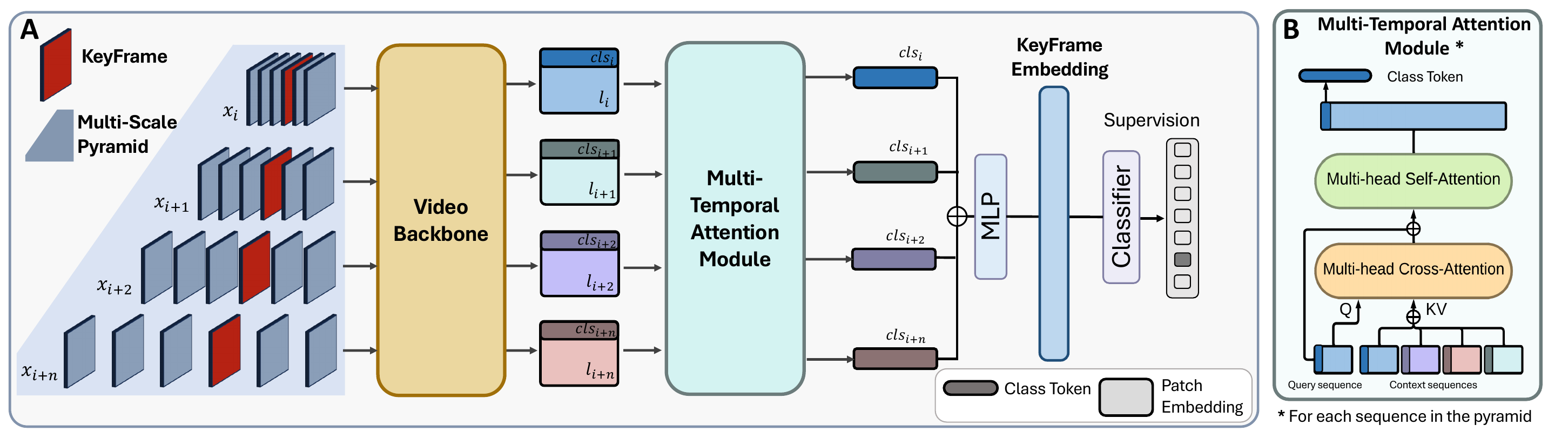}
    \caption{\textbf{MuST's Multi-Term Frame Encoder} utilizes MViT to capture embeddings of a temporal sequence sampled at multiple rates $(x_{i})$ with the keyframe placed at the middle position for offline setups and at the last position for online setups. Each sequence embedding $(l_{i})$, along with their corresponding class token $(cls_{i})$, goes through the Multi-Temporal Attention Module with the rest of the sequence embeddings of the pyramid as temporal context. A final rich multi-term embedding of the keyframe is generated by concatenating class tokens from each scale and processing them through a multi-layer perceptron (MLP). 
    }
    \label{fig:frame_encoder}
\end{figure}

As illustrated in Fig. \ref{fig:frame_encoder}, MuST begins by constructing a pyramid of sequences around each keyframe, capturing a hierarchy of temporal scales with increasing sample rates \cite{stergiou2023tempr}. Our method leverages a Video Backbone \cite{fan2021mvit} to obtain rich feature representations of the temporal context around a surgical keyframe provided by each sampled sequence. Then, our Multi-Temporal Attention Module captures interdependencies between the different temporal scales and retains the class token from each sequence. Furthermore, we concatenate the class tokens through a Multi-Layer Perceptron (MLP) to generate rich multi-term embeddings. Finally, our TCM self-attends the generated frame embeddings from the MLP and enhances coherence in predictions across an extensive temporal window, as shown in  Fig. \ref{fig:arch}.

\noindent \textbf{Temporal Multi-Scale Pyramid.} We sample sub-sequences from a surgical video at increasing rates to construct a temporal multi-scale pyramid. In offline setups, each sequence places the keyframe in the middle position, whereas in online setups, it is placed at the last position. Thus, this pyramid offers finer-grained temporal detail with lower temporal coverage at the first levels, while the last levels provide broader but sparse temporal context. This hierarchical structure enables simultaneous analysis at multiple temporal resolutions and captures information efficiently across various time scales.

\noindent Given a total number of \textit{N} sampling scales, we define the set of sampled sub-sequences as $\textit{X}={\{x_i\}}_i^N$ where each sequence $x_i \in \mathbb{R}^{T \times H \times W \times 3}$ has \textit{T} RGB frames of \textit{H×W} pixels and the temporal sampling stride of a sequence $x_i$ increases as $i$ approaches $N$. For each input sub-sequence, we use a shared video backbone $\mathcal{M}(\bullet)$ to compute a sequence of spatio-temporal embeddings $l_i \in \mathbb{R}^{T' \times D}$ with \textit{T'} embeddings of length \textit{D}. We define the set of output embeddings from the video backbone as $L={\{l_i\} }_i^N$ where $l_i=\mathcal{M}(x_i)$.

\noindent \textbf{Video Backbone.} We adopt MViT \cite{fan2021mvit} as our video backbone. MViT processes a fixed temporal window throughout multiple stages containing several Transformer blocks. At the beginning of each stage, a Multi-Head Pooling Attention (MHPA) mechanism reduces the space-time resolution while augmenting the feature dimension. Thus, MViT produces an implicit multi-scale feature pyramid where the initial stages capture detailed spatial resolutions while the final stages encode a shorter sequence of complex spatio-temporal embeddings. Additionally, MViT concatenates a learnable class token to capture a single embedding of the entire sequence. For further details, we refer the reader to the original MViT~\cite{fan2021mvit} paper. MuST retains the class token and the patch embeddings from the video backbone for each pyramid level.

\noindent  \textbf{Multi-Temporal Attention Module.} We introduce a novel Multi-Temporal Attention Module to compute relationships among the spatio-temporal representations extracted from each pyramid scale \cite{chen2021crossvit}, as detailed in  Fig. \ref{fig:frame_encoder}B. This module is composed of two units: a Multi-Temporal Cross-Attention (MTCA) unit to explore sequence inter-dependencies, denoted as $\textit{MTCA}(\bullet)$, and a residual Multi-Temporal Self-Attention (SA) unit to identify relationships among all computed embeddings, symbolized by $\textit{SA}(\bullet)$. The \textit{MTCA} module computes cross-attention between each embedding sequence with the concatenation of all sequences. For the \emph{i}-th set of embeddings generated from the \emph{i}-th sub-sequence, we compute $MTCA$'s attention as follows,

\setlength{\abovedisplayskip}{1pt}
\setlength{\belowdisplayskip}{1pt}
\begin{equation}
    l' = \text{concat}(L), \quad Q_i = {W_i}^{Q} l_i, \quad K_i = {W_i}^{K} l'_i, \quad V_i = {W_i}^{V} l_i'
\end{equation}

\begin{equation}
    \label{eq:cross_attention}
    \mathbf{MTCA}(l_i, l'_i) = \text{softmax}\left( \frac{Q_i \cdot {K_i}^{T}}{\sqrt{d_{k_i}}} \right) V_i
\end{equation}

\noindent Where $\textit{concat}(L)$ represents the concatenation operation across the sequence axis and $W_i^Q$, $W_i^K$, and $W_i^V$ represent the learnable weights of the queries \textit{(Q)}, keys \textit{(K)}, and values \textit{(V)} for the \emph{i}-th embedding, respectively. We designate the set of output embeddings from the $MTCA$ by $C={\{c_i\}}_i^N$ where  $c_i = \text{MTCA}(l_i, l'_i)$ and $c_i \in \mathbb{R}^{T' \times D}$. Further on, for each $c_i$ we calculate self-attention in the $SA$ module as follows:

\begin{equation}
    {c'_i} = \text{concat}({l_i}, C), \quad {Q_i} = {W_i^{Q}} {c_i'}, \quad {K_i} = {W_i^{K}} {c'_i}, \quad {V_i} = {W_i^{K}} {c'_i}
\end{equation}

\begin{equation}
    \mathbf{SA}(c'_i) = \text{softmax}\left( \frac{{Q_i} \cdot {K_i}^{T}}{\sqrt{d_{k_i}}} \right) {V_i}
\end{equation}

\noindent Where $concat({l_i}, C)$ represents the concatenation across the sequence dimension of all the embeddings in $C$ with the $l_i$ sequence produced by the video backbone as a residual connection, thus $c'_i \in \mathbb{R}^{(N+1)T'\times D}$. Finally, we concatenate all class tokens into a single embedding $p' \in \mathbb{R}^{ND}$ which is linearly transformed by an MLP into the final multi-term frame embedding denoted as $p \in \mathbb{R}^{ND}$ where $p = MLP(p')$. We train our shared video backbone and our Multi-Temporal Attention Module by adding a linear classifier that projects each $p$ into a class probability distribution corresponding to the phase class of the middle frame in the multi-sequence pyramid.

\noindent  \textbf{Temporal Consistency Module (TCM).} We adopt Transformers' self-attention to capture relationships between multiple frames and enhance their understanding of long-term dependencies. Given a video $V \in \mathbb{R}^{F\times H\times W \times 3}$ with $F$ frames, we construct overlapping windows $v={\{p_j\}}_j^{F'}$ of $F'$ frames ($F'<F$) where $p_j$ corresponds to the multi-term frame embedding generated for the \emph{j}-th frame in the window. Thus, the total number of windows in $v$ will be equal to $1 + \frac{F-F'}{F'-Overlap}$. We perform self-attention as follows:

\begin{equation}
    \mathbf{v'} = \text{concat}(v) + \text{PE}, \quad \text{Q} = W^Q v', \quad \text{K} = W^K v', \quad \text{V} = W^V v'
\end{equation}

\begin{equation}
    \text{TCM}(v) = \text{softmax}\left(\frac{Q \cdot K^{T}}{\sqrt{d_k}}\right) V
\label{equation:selfattn}
\end{equation}

\noindent Where $concat(v)$ is the concatenation of all embeddings in $v$ into a sequence of $F'$ embeddings, and $PE$ corresponds to a cosine positional embedding. Finally, we linearly project each output embedding into a phase class distribution. For offline inference, we average all the probability distributions obtained for each frame throughout all sampled overlapping windows. 

%% file: Sections/03_Experiments.tex
\section{Experiments and Results} \label{sec:experiments}

\subsection{Experimental Setup}

\textbf{Datasets.} We evaluate the performance of MuST in three surgical datasets: HeiChole \cite{wagner2023comparative},  GraSP \cite{ayobi2024pixelwise}, an extended version of PSI-AVA \cite{valderrama2022towards}, and MISAW \cite{huaulme2021micro}. These datasets are standard benchmarks in surgical workflow analysis from different domains. We adhere to each dataset's public benchmarks and metrics to ensure fair comparison with other models. Thus, we evaluate using the F1-score in HeiChole in an online setup and mean Average Precision (mAP) in GraSP and MISAW offline. Since HeiChole's test data is not public, we used the training and validation sets proposed by \cite{wagner2023comparative} for comparison with other methods. For GraSP, we perform ablation studies using its cross-validation set and report final results using its test set.

\textbf{Implementation Details.} We initialize the weights from an MViT-B model pretrained on phase recognition tasks for 20 epochs. Using these pretrained weights, we train the MuST Multi-Term Encoder for an additional five epochs, utilizing three NVIDIA Quadro RTX 8000 GPUs with a batch size of 18. For the offline setup, we construct the Temporal Multi-Scale Pyramid using four sequences of 16 frames each, with sampling rates of 1, 4, 8, and 12 seconds. We use 24 frames with the same sampling rates for the online setup. Specifically for HeiChole, we adapt the temporal window to include only past frames to ensure a fair comparison with other online methods. We train the Temporal Consistency Module (TCM) for 20 epochs on a single GPU with a batch size of 256. The temporal window covers 10\% of the mean video duration for offline datasets and 5\% for online datasets, with a 90\% overlap for all datasets. The training for all modules in MuST utilizes cross-entropy loss, optimized using AdamW with a cosine learning rate scheduler, a weight decay of $1 \times 10^{-4}$, and an initial learning rate of $1 \times 10^{-4}$. Our model comprises 68 million parameters.

\subsection{Experimental Validation}

\begin{figure}[]
    \centering
    \includegraphics[width=0.8\linewidth]{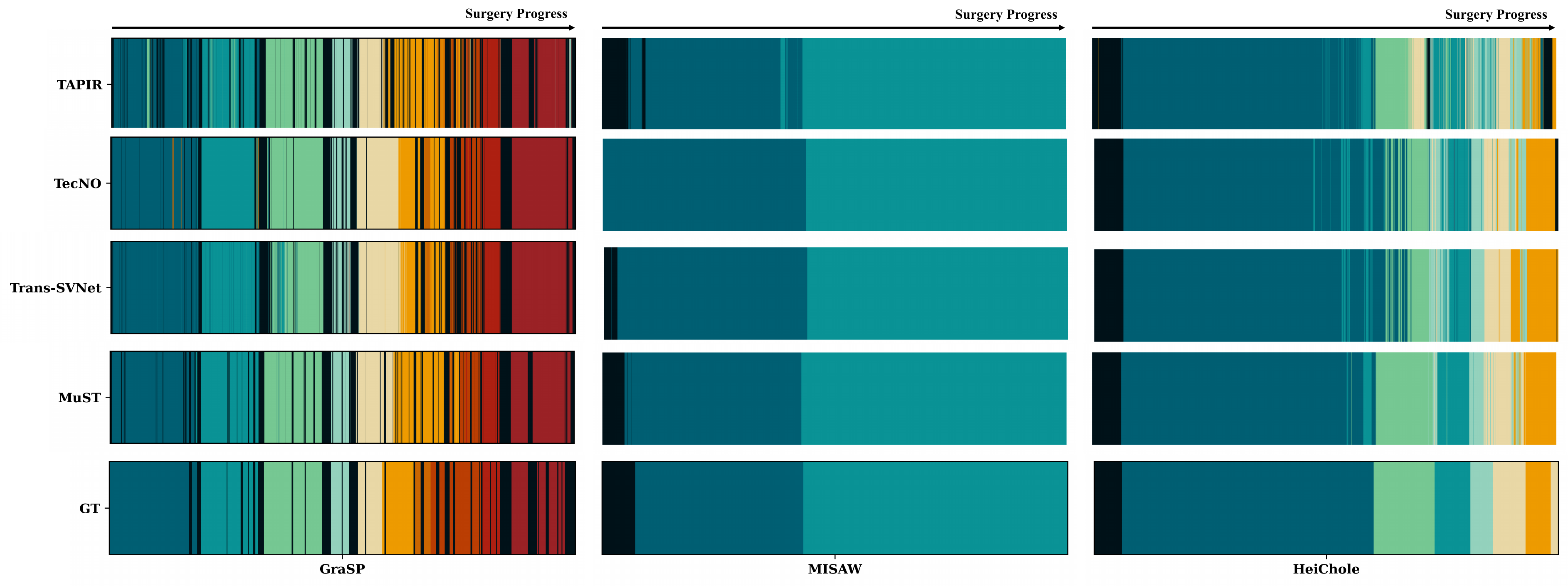}
    \caption{\textbf{Qualitative results.} Qualitative comparisons with state-of-the-art methods on one representative video from each dataset. }
    \label{fig:qualitative_results}
\end{figure}

We show the overall performance in Table \ref{tab:datasets-results} and our qualitative comparisons in Fig. \ref{fig:qualitative_results}. We compare our model with open-source, state-of-the-art methods. We train and optimize these methods on our selected benchmarks to ensure fair and accurate comparisons. MuST outperforms previous state-of-the-art methods across all datasets, achieving excellent results in both online and offline setups. These results highlight the effectiveness of multi-scale temporal reasoning in ensuring consistent predictions across various temporal intervals.

\begin{table}[]
\centering
\caption{\textbf{Comparative Results of MuST in different benchmarks.} The best results are shown in bold. All methods were adapted to conduct offline or online inference according to the dataset.}
\label{tab:datasets-results}
\resizebox{\textwidth}{!}{%
\begin{tabular}{cccccccccc}
\cline{1-2} \cline{5-6} \cline{9-10}
\multicolumn{2}{c}{a)   Grasp}        &           &           & \multicolumn{2}{c}{b) MISAW}          &           &           & \multicolumn{2}{c}{c) HeiChole}       \\ \cline{1-2} \cline{5-6} \cline{9-10} 
Model                & Phases mAP     &           &           & Model                & Phases mAP     &           &           & Model                & F1-score       \\ \cline{1-2} \cline{5-6} \cline{9-10} 
TAPIS                & 76.07          &           &           & TAPIS                & 97.14          &           &           & TAPIS                & 73.41          \\
TeCNO                & 77.10          &           &           & TeCNO                & 95.58          &           &           & TeCNO                & 69.35          \\
Trans-SVNet          & 76.54          &           &           & Trans-SVNet          & 90.38          &           &           & Trans-SVNet          & 71.85          \\ \cline{1-2} \cline{5-6} \cline{9-10} 
\textbf{MuST (ours)} & \textbf{79.14} & \textbf{} & \textbf{} & \textbf{MuST (ours)} & \textbf{98.08} & \textbf{} & \textbf{} & \textbf{MuST (ours)} & \textbf{77.25}
\end{tabular}%
}
\end{table}

Our method consistently overperforms TAPIS, which employs an MViT backbone with a fixed temporal window. Compared to TeCNO, MuST improves phase recognition metrics in short- and medium-duration time intervals. We present per-class performance and phase mean durations across all datasets in Table 1 of the Supplementary Material. These results prove the advantages of incorporating spatio-temporal reasoning of sequences facilitated by MuST’s backbone instead of per-frame embeddings coming from a CNN classifier. Along with the Multi-Temporal Attention Module and the TCM, MuST distinguishes phase time intervals and durations more accurately.

Finally, compared to Trans-SVNet, which relies on short-term fixed-size windows and per-frame prediction, MuST effectively captures rapid phase changes, particularly evident in the GraSP dataset (Fig. \ref{fig:qualitative_results}), where phases exhibit highly variable time durations. This capability arises from the multi-term reasoning from the MTFE. This encoder learns relationships within each sub-sequence in the pyramid, producing rich multi-term frame embeddings that are subsequently refined by the TCM for segment predictions that relate a wider temporal context within the sequence. Furthermore, we conducted an additional validation of MuST on the Cholec80 dataset \cite{twinanda2016endonet} and obtained an F1-score of 85.57\%, precision of 84.6\%, and recall of 86.6\%, achieving a highly competitive performance, comparable to current state-of-the-art methods on this benchmark \cite{liu2023skit,liu2023lovit}. 

\subsection{Ablation Experiments}

We conduct ablations on the cross-validation set of the GraSP benchmark, as detailed in Table \ref{table:ablations} and displayed in Fig. 3 of the Supplementary Material. Our experiments demonstrate that using multiple sequences sampled at the same rate as input leads to a performance increase, likely due to higher redundancy that allows a wider understanding of the patterns present in the data. Similarly, we observe a significant performance boost when incorporating the Temporal Multi-Scale Pyramid, which constructs windows at varying sample rates. This approach aggregates information from multiple scales, resulting in richer contextual information. Adding cross-attention and self-attention mechanisms further improves the model's performance as they establish cross-scale reasoning that enables interactions between sequences in the pyramid. This is evident in the cross-attention maps in Fig. 2 of the Supplementary Material. Lastly, the TCM significantly enhances the model's prediction coherence by incorporating long-term dependencies between frame embeddings through its transformer encoder architecture. This improvement is illustrated in Fig. 3 of the Supplementary Material.

\begin{table}[]
\centering

\caption{\textbf{Ablation experiments results for MuST in the GraSP benchmark.}
The Multi-Sequence column indicates if the model processes multiple input windows, and the Temporal Multi-Scale Pyramid column denotes the use of multiple temporal scales. 
The Multi-Temporal Attention Module column indicates the use of attention mechanisms across multiple temporal scales. 
TCM represents the inclusion of the Temporal Consistency Module.  A checkmark (\checkmark) indicates the presence of an attribute, while a blank space denotes its absence. The best result is highlighted in bold.}

\resizebox{10.5cm}{!}{
\begin{tabular}{c|c|c|c|c}
\hline
\rowcolor[HTML]{FFFFFF} 
\textbf{Multi-Sequence} & \textbf{\begin{tabular}[c]{@{}c@{}}Temporal Multi-Scale \\ Pyramid\end{tabular}} & \textbf{\begin{tabular}[c]{@{}c@{}}Multi-Temporal \\ Attention Module\end{tabular}} & \textbf{TCM} & \textbf{mAP} \\ \hline
\rowcolor[HTML]{F8F8F8} 
 &  &  &  &   72.57 $\pm$ 0.09   \\
 \rowcolor[HTML]{FFFFFF} 
$\checkmark$ &  &  &  &   73.38 $\pm$ 2.30   \\
\rowcolor[HTML]{F8F8F8} 
$\checkmark$ & $\checkmark$ &  &  &   75.87 $\pm$ 2.19   \\
\rowcolor[HTML]{FFFFFF} 
$\checkmark$ & $\checkmark$ & $\checkmark$ &  &   76.13 $\pm$ 1.74   \\
\rowcolor[HTML]{F8F8F8} 
$\checkmark$ & $\checkmark$ & $\checkmark$ & $\checkmark$ & \textbf{77.34 $\pm$ 1.02}   \\ \hline
\end{tabular} }
\label{table:ablations}
\end{table}

%% file: Sections/04_Conclusions.tex
\section{Conclusions} \label{sec:conclusions}

In this work, we introduce a novel method for surgical phase recognition that effectively addresses the challenges associated with varying phase durations in surgical data. By leveraging a temporal pyramid and a cross-attention module, we enrich the temporal context and facilitate multi-scale learning, capturing short- and mid-term dependencies. Additionally, we present a Temporal Consistency Module to enhance long-term reasoning and strengthen the model's overall performance. We demonstrate that MuST outperforms previous state-of-the-art methods across different surgical datasets and establishes a flexible framework for future research in complex temporal modeling in surgical settings.